\documentclass[10pt,twocolumn,letterpaper]{article}

\usepackage{iccv}
\usepackage{url}            % simple URL typesetting
\usepackage{booktabs}       % professional-quality tables
\usepackage{amsfonts}       % blackboard math symbols
\usepackage{nicefrac}       % compact symbols for 1/2, etc.
\usepackage{microtype}      % microtypography
\usepackage{xcolor}        % colors
\usepackage{times}
\usepackage{tabu}
\usepackage{epsfig}
\usepackage{graphicx}
\usepackage{dsfont}
\usepackage{multirow}
\usepackage{enumerate}
\usepackage{enumitem}
\usepackage{pifont}
\usepackage{comment}
\usepackage{caption}
\usepackage{silence}
\usepackage{physics}
\usepackage{nicematrix}
\usepackage{wrapfig}
\usepackage{float}
\usepackage{tikz}
\usepackage{bm}
\usepackage{bbm}
\usepackage{makecell}
\usepackage{soul}

\usepackage{tcolorbox}

\usepackage{colortbl}
\definecolor{mygray}{gray}{0.9}

\def\halfcheckmark{\textcolor{black}{\ding{52}}{\small\textcolor{black}{\kern-0.7em\ding{55}}}}

\usepackage{xspace}
\def\ourtask{IDG\xspace}
\def\ourdataset{DiffGround\xspace}
\def\ourmodel{DiffTracker\xspace}

\definecolor{iccvblue}{rgb}{0.21,0.49,0.74}
\usepackage[pagebackref,breaklinks,colorlinks,allcolors=iccvblue]{hyperref}

\title{Image Difference Grounding with Natural Language}

\author{
  Wenxuan Wang$^{1,2,3}$\thanks{Equal technical contribution.} \quad
  Zijia Zhao$^{1,2*}$\quad
  Yisi Zhang$^{4*}$\quad
  Yepeng Tang$^{5}$\quad
  Erdong Hu$^{1,2}$\quad \\
  Xinlong Wang$^3$\quad
  Jing Liu$^{1,2}\thanks{Corresponding author.}$\\
  {$^1$
  Institute of Automation, Chinese Academy of Sciences}\\
  {$^2$ School of Artificial Intelligence, University of Chinese Academy of Sciences}\\
  {$^3$ Beijing Academy of Artificial Intelligence}
  {$^4$ University of Science and Technology Beijing} \\
  {$^5$ Institute of Information Science, Beijing Jiaotong University} \\
  \tt\small \{wangwenxuan2023@ia.ac.cn, wangxinlong@baai.ac.cn, jliu@nlpr.ia.ac.cn\}
}

\begin{document}

\twocolumn[{
\renewcommand\twocolumn[1][]{#1}
\maketitle 
\vspace{-8mm}
\begin{center} 
\centering 
\includegraphics[width=0.96\textwidth]{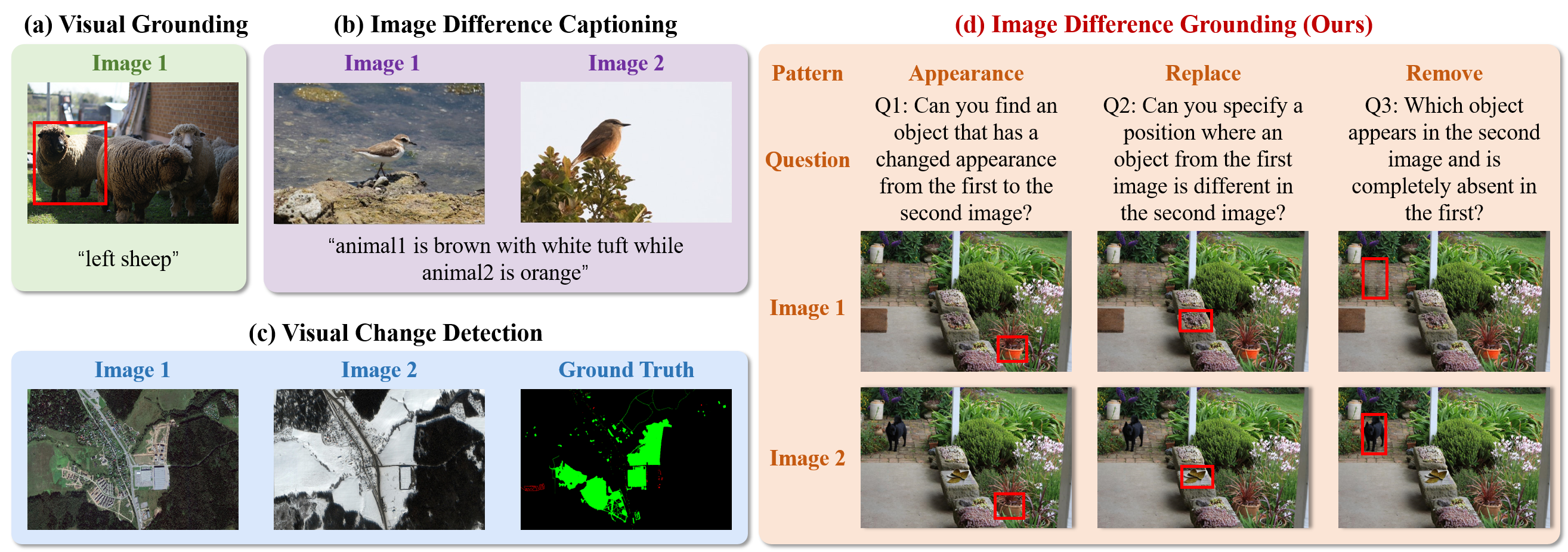} 
\vspace{-2mm}
\captionof{figure}{
Task Comparison between classic visual grounding (VG), visual change detection (VCD), image difference captioning (IDC),  and our proposed image difference grounding (IDG) for finer-grained vision-language understanding.
}
\label{fig:intro} 
\end{center}
}]

\renewcommand{\thefootnote}{\fnsymbol{footnote}}
\footnotetext[1]{Equal contribution.}
\footnotetext[2]{Corresponding author.}

\begin{abstract}
Visual grounding (VG) typically focuses on locating regions of interest within an image using natural language, and most existing VG methods are limited to single-image interpretations. 
This limits their applicability in real-world scenarios like automatic surveillance, where detecting subtle but meaningful visual differences across multiple images is crucial. 
Besides, previous work on image difference understanding (IDU) has either focused on detecting all change regions without cross-modal text guidance, or on providing coarse-grained descriptions of differences.
Therefore, to push towards finer-grained vision-language perception, we propose Image Difference Grounding (\ourtask), a task designed to precisely localize visual differences based on user instructions.
We introduce \ourdataset, a large-scale and high-quality dataset for \ourtask, containing image pairs with diverse visual variations along with instructions querying fine-grained differences. 
Besides, we present a baseline model for \ourtask, \ourmodel, which effectively integrates feature differential enhancement and common suppression to precisely locate differences. 
Experiments on the \ourdataset dataset highlight the importance of our \ourtask dataset in enabling finer-grained IDU. 
To foster future research, both \ourdataset data and \ourmodel model will be publicly released.

\end{abstract}

\section{Introduction}
\label{sec:introduction}

The rapid advancement of multimodal embodied intelligence \cite{ahn2022can, reed2022generalist, driess2023palm, shah2023lm, gao2023physically, o2023open, brohan2023rt, huang2023embodied} has underscored the importance of visual grounding (VG), which enables agents to locate and interact with objects based on natural language in open-world environments. 
Unlike traditional vision tasks focused on predefined object categories, VG requires recognizing objects in dynamic, real-world contexts, which makes it a crucial task for developing intelligent systems that can comprehend natural language and visual information. 
As the demand for more autonomous, context-aware systems grows, VG plays a key role in advancing human-machine interactions and bridging language with vision, offering significant potential for applications in open-world scene understanding and language-based human-object interaction.

Since the introduction of VG in \cite{mao2016generation}, it has become a central task in vision-language (V-L) understanding, garnering significant research attention. 
Numerous multimodal frameworks have been designed to tackle the challenges of feature extraction and alignment between V-L modalities \cite{hu2016segmentation, deng2021transvg, wang2022cris, yang2022lavt, wang2024cm, wang2024unveiling}. 
However, most existing approaches focus on grounding within a single image, limiting their ability to detect and analyze visual differences across multiple images. 
This ability is essential for real-world applications, like automatic surveillance and monitoring, where agents must compare visual information from different frames or sources. 
Furthermore, current benchmark datasets, while effective for single-image grounding, lack the resources necessary to enable models to perceive cross-image visual discrepancies. 
In summary, while detecting visual discrepancies between images is crucial, existing methods and datasets in the VG field do not support such advanced capability, creating a substantial gap in the research. 
\textit{Thus, exploring how to transcend VG beyond single-image constraints to handle cross-image differences is key to advancing intelligent systems, which is the primary focus of this work.}

Prior to this work, several studies have made significant progress in understanding image differences, generally categorized into unimodal visual change detection and V-L based image difference captioning (IDC). 
Traditional visual change detection (VCD), aimed at identifying visual differences between two images, has been widely studied \cite{lei2023lightweight,yan2023transy,fang2023changer,luo2023multiscale,feng2023change,ding2024joint,chen2024changemamba,li2024new}. 
However, this task is unimodal, focusing on detecting image differences solely based on visual input, and lacks a deep connection between visual discrepancies and textual descriptions. 
In contrast, V-L based IDC task has gained great attention \cite{jhamtani2018learning,park2019robust,shi2020finding,huang2021image,hosseinzadeh2021image,kim2021agnostic,guo2022clip4idc,sun2022bidirectional,yao2022image,tu2023adaptive,tu2023neighborhood,tu2023self}, driven by the need for cross-image difference captioning. 
However, IDC primarily offers global descriptions of image differences and does not enable fine-grained localization of specific difference areas as per user requests. 
In summary, while text-guided cross-modal perception and fine-grained V-L understanding are essential for multimodal systems in real-world scenes, few studies have addressed these aspects simultaneously or explored their connection. 
This creates a significant research gap on fine-grained image difference grounding driven by V-L information,which is crucial in terms of two aspects.
First, users typically expect agents to identify specific cross-image differences based on descriptive instructions, rather than all the differences (mostly uninterested) at once, which highlights the need for multimodal agents to accurately locate regions of interest. 
Second, fine-grained cross-image difference perception can positively promote visual perceiving within a single image, particularly for subtle visual discrepancies, thereby advancing classic VG task.

Therefore, in this work, we attempt to fill the aforementioned critical gap that has been previously neglected, advancing beyond classic single-image VG and coarse-grained IDC towards fine-grained cross-image difference grounding. 
As illustrated in Fig. \ref{fig:intro}, we introduce a new V-L understanding task, Image Difference Grounding (\ourtask), and create a high-quality benchmark dataset, \ourdataset, using advanced image editing techniques and manual annotations. 
Specifically, \ourdataset contains image pairs with diverse visual discrepancies (\ie, object appearance changes, removal, and replacement) annotated with textual queries. 
Building on these fine-grained annotations, we propose \ourmodel, a new multimodal framework designed to precisely realize \ourtask. 
By incorporating our well-designed Difference Enhancement Module (DEM), which progressively enhances unique visual differences and suppresses shared features at each stage of the vision backbone, \ourmodel ensures high precision in grounding cross-image differences.
Extensive evaluations on \ourdataset demonstrate that \ourmodel outperforms existing VG models, showcasing its superior ability to localize image differences in response to user instructions. 
By releasing both \ourdataset and \ourmodel, we aim to provide essential resources for advancing fine-grained \ourtask and promoting more capable multimodal embodied systems.

Our main contributions can be summarized as follows:
\setlist{nolistsep}
\begin{itemize}[noitemsep,leftmargin=*]
    \item 
    We propose a new \ourtask task, which requires accurate localization of visual differences between image pairs via language expressions, pushing for finer-grained cross-image difference perception beyond classic VG and IDC.
    \item 
    We introduce \ourdataset, the first large-scale benchmark dataset designed for \ourtask task, containing high-quality image pairs with diverse visual variations and textual instructions to capture fine-grained differences.
    \item 
    We develop a new multimodal framework namely \ourmodel and introduce the Difference Enhancement Module for \ourtask task. Experiments on \ourdataset dataset demonstrate \ourmodel's superiority in handling \ourtask compared with existing VG methods.
\end{itemize}

\begin{figure*}[htbp]
    \centering
    \includegraphics[width=0.86\textwidth]{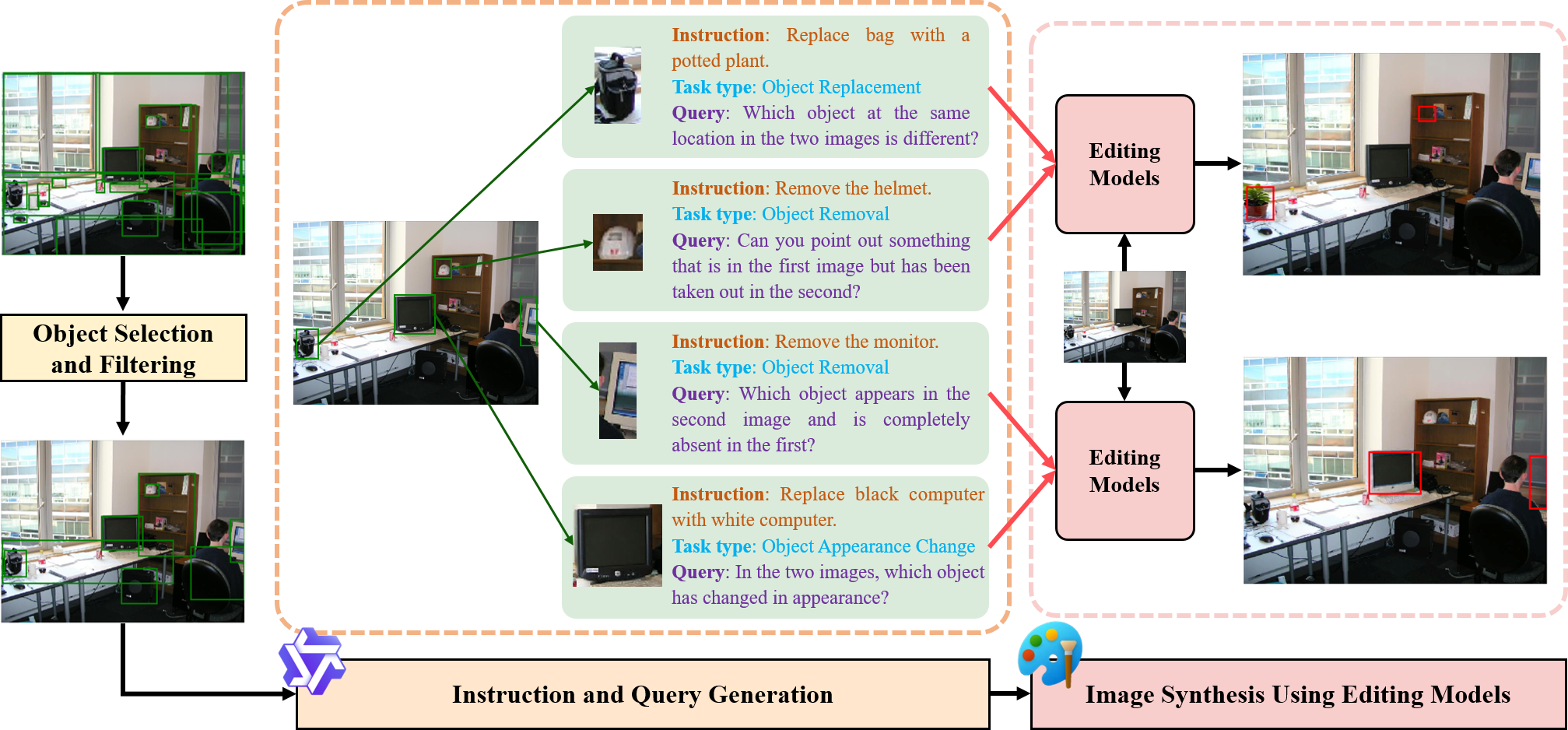}
    \vspace{-8pt}
    \caption{
    The illustration of our data engine for building the \ourdataset dataset. 
    }
    \label{fig_dataengine}
    \vspace{-12pt}
\end{figure*}

\section{Related Work} 
\label{sec:related_work}

\paragraph{Visual Grounding}\!\!\!\!\!aims to precisely locate the target object corresponding to a given language expression within an image. As the two fundamental VG tasks differing by their output form, Referring Expression Comprehension (REC) \cite{mao2016generation,chen2018real,deng2021transvg,deng2023transvg++,bai2023qwen,zhu2023minigpt,chen2023minigpt,chen2023shikra,you2023ferret,wang2024qwen2} and Referring Expression Segmentation (RES) \cite{hu2016segmentation,ye2019cross,ding2021vision,yang2022lavt,wang2022cris,zou2023segment,zou2023generalized,lai2024lisa,xia2024gsva,wang2024cm,wang2024unveiling,rasheed2024glamm} have been extensively studied, with REC being the primary focus of this work due to its importance. 
However, existing VG approaches typically ground objects within a single image, limiting their ability to handle cross-image visual content. 
In this work, we aim to push the boundaries of VG by addressing fine-grained, cross-image differences, a crucial capability for real-world applications that require comparing visual information across multiple images.

\vspace{-12pt}
\paragraph{Image Difference Understanding}\!\!\!\!\!
focuses on analyzing visual discrepancies between two given images, a crucial task for applications requiring cross-image perception and comparison. 
Early works in this area primarily concentrate on VCD, aiming to identify all visual differences between two images. 
These unimodal methods, such as \cite{lei2023lightweight,yan2023transy,fang2023changer,luo2023multiscale,feng2023change,ding2024joint,chen2024changemamba,li2024new}, detect all the discrepancy areas based solely on visual information, without incorporating textual input, limiting their ability to localize fine-grained differences as per user instructions. 
In contrast, IDC has gained significant attention recently by shifting focus toward V-L perception, where models are required to generate textual descriptions of image differences \cite{jhamtani2018learning, park2019robust, shi2020finding, huang2021image, hosseinzadeh2021image, kim2021agnostic, guo2022clip4idc, sun2022bidirectional, yao2022image, tu2023adaptive, tu2023neighborhood, tu2023self}. 
However, existing IDC methods generally focus on coarse-grained difference captioning, struggling to identify and localize fine-grained visual differences. 
In this work, we attempt to address this gap by advancing image difference understanding to enable precise localization of cross-image differences, an essential capability for real-world applications requiring detailed comparisons based on user requests.

\section{IDG Task \& DiffGround Dataset} 
\label{sec:task_benchmark}

\subsection{IDG Task Description}
\label{subsec:task_definition}

The proposed IDG task aims at localizing a specific cross-image difference region based on a user-provided textual query.
It requires two key capabilities: fine-grained localization to detect subtle image discrepancies, and text-image interaction to interpret the query and locate the described difference. 
IDG differs from previous IDC and VCD tasks in two main ways: first, it requires more precise localization of differences with bounding box output, and second, it integrates a text-guided query, requiring strong multimodal interaction to identify differences based on language cues.

We formally define the IDG task as follows: given two similar images, \(I_1\) and \(I_2\), along with a textual query \(T\), the goal is to return the location of the difference referenced by query, expressed as a bounding box \((x_1, y_1, x_2, y_2)\).
We categorize the IDG task into three sub-tasks based on the difference type of query: \textit{appearance}, \textit{replace}, and \textit{remove}. 
In both \textit{appearance} and \textit{replace}, the image difference reflects a change in an object at the same location between the two images. 
\textit{Replace} involves coarse-grained variations, where the object category changes, and \textit{appearance} refers to fine-grained changes, where the object's appearance differs but its category remains the same.
As for \textit{remove}, the difference lies in the presence or absence of an object at a specific region. IDG task of \textit{remove} pattern is further divided into two cases: (1) the target is present in \(I_1\) but absent in \(I_2\), or (2) the object is present in \(I_2\) but missing in \(I_1\).

It is worth noting that we exclude potential object category information from the text query \(T\) to prevent the model from directly identifying target regions based on explicit category.
Besides, the paired images \(I_1\) and \(I_2\) contain multiple distinct differences, but only one of these differences corresponds uniquely to the provided text query \(T\). This ensures that the model cannot simply rely on detecting general image differences to produce the target bounding box, thereby requiring it to perform precise, query-guided localization.

\begin{figure*}[htbp]
    \centering
    \includegraphics[width=0.86\textwidth]{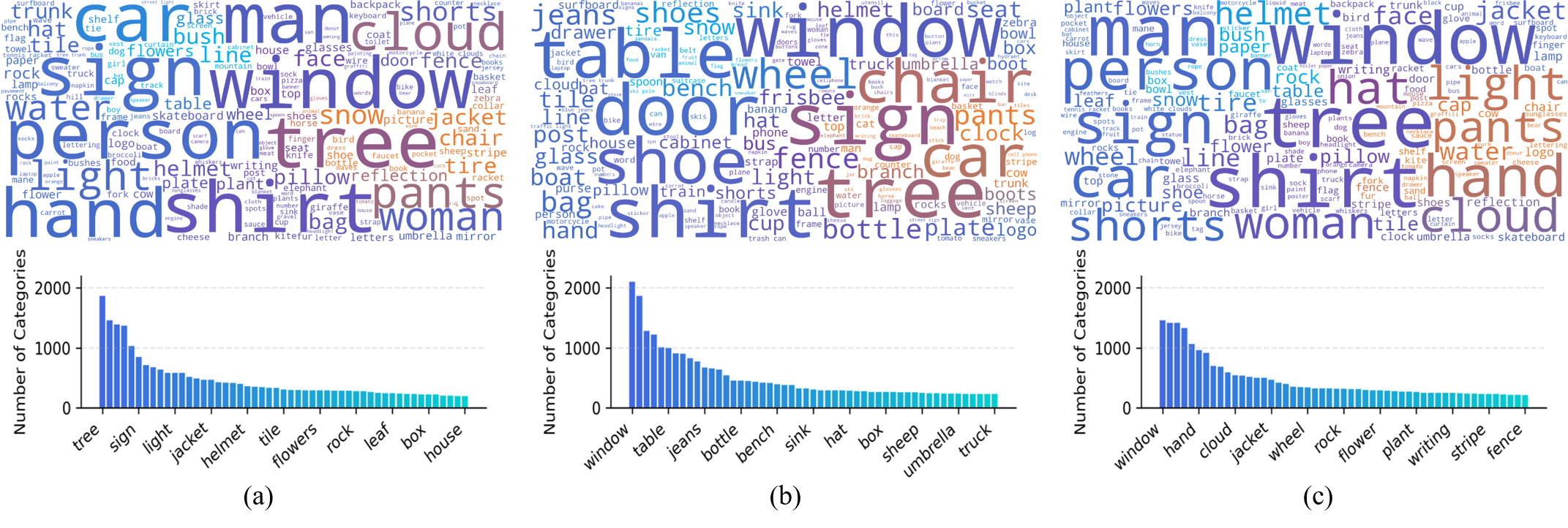}
    \vspace{-12pt}
    \caption{Our \ourdataset dataset statistics for each difference pattern. 
    (a), (b) and (c) respective illustrates the statistics of the word diversity for each object category (\ie, first row) and the occurrence frequency of different categories (\ie, second row) within a single difference pattern (\ie, coarse-grained object category change, fine-grained object appearance change and direct object removal).
    The horizontal coordinates for (a), (b), (c) are  the examples of the specific categories with the ranked top 50 highest vertical values.}
    \label{fig_data_analysis2}
    \vspace{-10pt}
\end{figure*}

\subsection{Data Collection Engine}
\label{subsec:data_engine}
In natural scenes, it is challenging to obtain large quantities of fine-grained image difference pairs, so we adopt a synthetic approach to construct our IDG task dataset. 
We utilize image editing models to create difference image pairs, with two primary types of edits: localized modifications, corresponding to the \textit{replace} and \textit{appearance} patterns, and localized removals, corresponding to the \textit{remove} category.
The Visual Genome (VG) dataset~\cite{krishna2017visual} serves as our grounding data source due to its extensive coverage of real-world scenes and its rich textual annotations for object categories and attributes. As shown in Fig. \ref{fig_dataengine}, the data collection pipeline for our DiffGround dataset involves three steps:

\textbf{Step 1: Object Selection and Filtering}: We first retrieve the bounding box coordinates \((x_1, y_1, x_2, y_2)\) and object categories from VG, then filter the objects based on the following criteria: (a) remove objects occupying less than 1\% or more than 20\% of the total image area; (b) exclude objects deemed difficult to modify according to predefined rules (\eg, parts or background); (c) eliminate objects with significant overlap (IoU $>$ 0.1) with others. For (b), we use the Qwen2-7B~\cite{yang2024qwen2} model to evaluate object replaceability, outputting a score from 0 to 10 based on the object category and prompt. Prompt details are in the \textcolor{blue}{Appendix}.

\textbf{Step 2: Instruction and Query Generation}: Then we generate editing instructions and queries for the filtered objects, ranked by their scores from step 1. 
If fewer than four objects having scores above 3, the image is excluded. 
We select four objects with scores above 3 and apply the following modifications: appearance change, category change, and two removals, ensuring each image pair contains all four types of differences. 
For appearance modification, we use the Qwen2-VL-7B~\cite{wang2024qwen2} model to extract common attributes and the Qwen2-7B~\cite{yang2024qwen2} model to generate alternatives. 
For category changes, the Qwen2-7B~\cite{yang2024qwen2} model is prompted to suggest a replacement category. 
For instruction generation, we insert the source object label \texttt{SRC} and modified attributes/categories \texttt{TGT} into a predefined template "Replace the [\texttt{SRC}] with [\texttt{TGT}]". Details on the prompt setup are in the \textcolor{blue}{Appendix}. For removal tasks, we use a removal template with the object category \texttt{SRC}, like "Remove the [\texttt{SRC}]". 
Next, we assign one of 80 predefined textual queries to each task. Note that these 80 candidate queries focus only on difference types, such as appearance changes or object removals, rather than providing object annotations. They require identifying the differences between the two images and understanding the characteristics of each.

\textbf{Step 3: Image Synthesis With Editing}: We then use the UltraEdit model~\cite{zhao2024ultraedit} for localized modifications and the Inpaint-Anything model~\cite{yu2023inpaint} for removals. After step 2, each VG image has two modification and two removal instructions. 
One modification and one removal are randomly applied in two iterative steps to create \(I_1\). 
The remaining instructions are applied to generate \(I_2\). 

Thus, each image pair \(I_1\) \& \(I_2\) contains two modified areas, corresponding to four distinct queries, each linked to a bounding box. 
This ensures that each query refers to a specific difference region, enabling precise localization. 
The final data format includes the image pair \(I_1\) \& \(I_2\) and four query instructions \([T_1, T_2, T_3, T_4]\), representing \textit{replace}, \textit{appearance}, and \textit{remove} in different images.

\subsection{DiffGround Dataset Details}
\label{subsec:dataset_details}

As shown in Table \ref{tab:dataset_details}, previous datasets faces challenge of lacked diversity and scale. 
They mainly focus on simpler tasks with less complex input structures and do not integrate V-L understanding effectively. 
In contrast, our DiffGround dataset offers a larger, more varied collection of images, supports dual-image and multimodal annotations, and demands fine-grained spatial reasoning, providing a more comprehensive evaluation of V-L perception models.

\begin{table}[htbp]
    \centering
    \small
    \caption{Comparison with classic VG \cite{kazemzadeh2014referitgame,yu2016modeling,nagaraja2016modeling}, VCD \cite{chen2020spatial,ji2018fully,shi2021deeply} and IDC \cite{jhamtani2018learning,park2019robust,tan2019expressing,forbes2019neural} datasets. 
    \# denotes the specific number, 
    where Expression, Cats and Avg Len denote the textual annotation types, object categories and average length of language expressions.
    ``-'' denotes corresponding properties are unavailable.
    }
    \vspace{-2mm}
    \setlength{\tabcolsep}{0.3pt} 
    \begin{tabular}{lccccc}
    \specialrule{.1em}{.05em}{.05em}
        Datasets & \#Imgs & \#Labels & Expressions & \#Cats & \#Avg Len \\
        \midrule
        \multicolumn{6}{l}{\color{gray}Classic Visual Grounding} \\
        ReferIt & 20K & 97K & Phrase & 238 & 3.2 \\
        RefCOCO & 20K & 50K & Phrase & 80 & 3.6 \\
        RefCOCO+ & 20K & 49K & Phrase & 80 & 3.5 \\
        RefCOCOg & 26K & 54K & Phrase & 80 & 8.4 \\
        \midrule
        \multicolumn{6}{l}{\color{gray}Visual Change Detection} \\
        LEVIR-CD & 637 & 23K & - & - & - \\
        LEVIR-CD+ & 985 & 35K & - & - & -  \\
        WHU-CD & 8K & 8K & - & - & -  \\
        SYSU-CD & 20K & 20K & - & - & -  \\
        \midrule
        \multicolumn{6}{l}{\color{gray}Image Difference Captioning} \\
        Spot-the-Diff & 13K & 13K & Caption & - & 11.0 \\
        CLEVR-Change & 40K & 80K & Caption & - & 8.2 \\
        Image-Editing-Request & 4K & 6K & Caption & - & 7.5 \\
        Birds-to-Words & 3K & 16K & Caption & - & \textbf{32.1} \\
        \midrule
        \multicolumn{6}{l}{\color{gray}Image Difference Grounding} \\
        \rowcolor{mygray}
        \textbf{\ourdataset} & \textbf{61K} & \textbf{244K} & \textbf{Instruction} & \textbf{13625} & 15.3 \\
    \specialrule{.1em}{.05em}{.05em}
    \end{tabular}
    \label{tab:dataset_details}
    \vspace{-12pt}
\end{table}

\begin{figure}[htbp]
    \centering
    \includegraphics[width=0.45\textwidth]{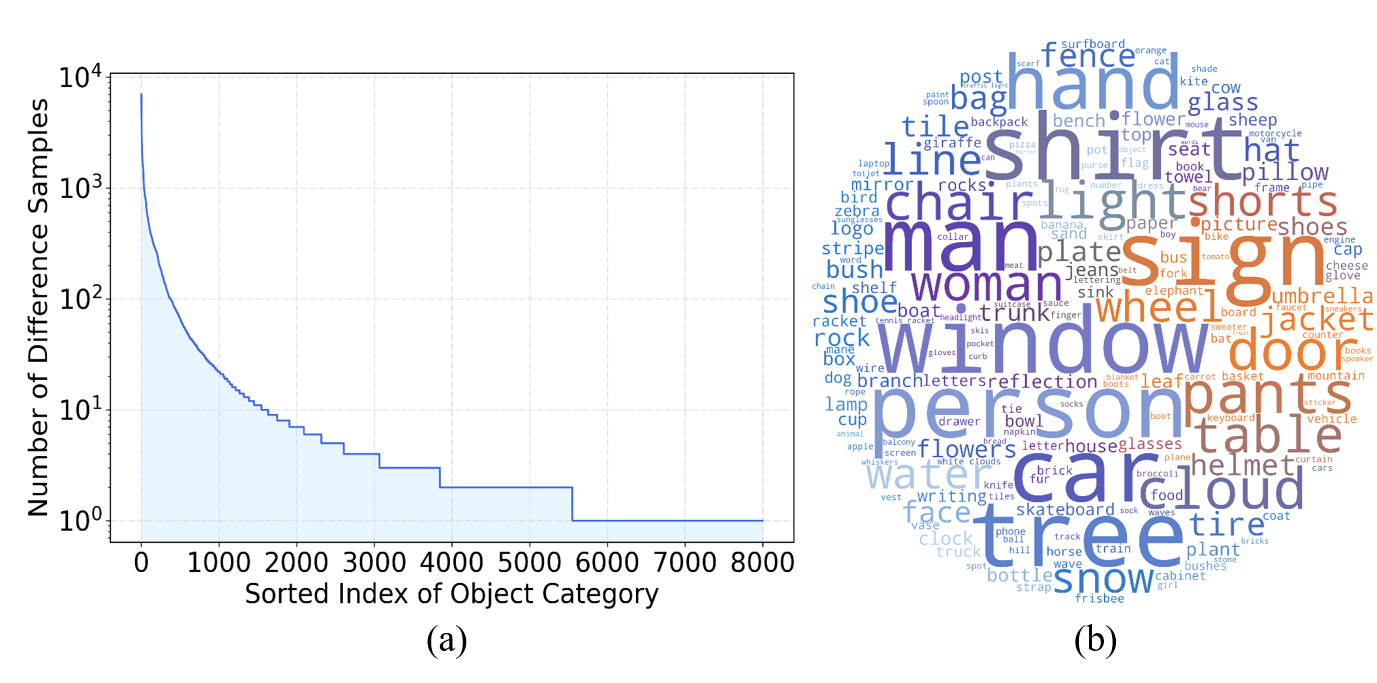}
    \vspace{-12pt}
    \caption{\ourdataset dataset statistics.
    (a) the number of IDC instructions per object's category (\ie, ranked top 8K) in the log scale. 
    (b) the word cloud highlights the head categories.
    }
    \label{fig_data_analysis1}
    \vspace{-12pt}
\end{figure}

\textbf{Rich Data Variety.}
Our DiffGround contains 61k images and 244k objects, greatly larger than previous VG and IDC datasets. 
The expanded size of both training and testing data enhances model optimization and enables more comprehensive evaluation. 
Additionally, DiffGround includes 13,625 distinct object categories, offering a broader variety of target types as shown in \cref{fig_data_analysis1}. This diversity surpasses that of prior datasets across multiple tasks, making DiffGround more representative of open-world scenarios.

\textbf{Dual-Image Perception.}
Compared to traditional VG task, IDG is more challenging and involves analyzing two images simultaneously to identify their differences, aligning closely with human interaction with visual information everyday and demanding more complex scene understanding. 
Compared with classic VG benchmarks~\cite{kazemzadeh2014referitgame,yu2016modeling,nagaraja2016modeling},
our DiffGround supports the IDG task by providing high-quality, V-L annotated data, enabling models to effectively perceive image differences across multiple inputs.

\textbf{Multimodal Reasoning.}
In contrast to conventional change detection datasets~\cite{chen2020spatial,ji2018fully,shi2021deeply}, our DiffGround dataset incorporates textual guidance, requiring the model to engage in multimodal interactions. Unlike traditional VG datasets, the text queries in DiffGround do not include explicit object labels, instead being more implicit and requiring model's advanced V-L reasoning capabilities.

\textbf{Fine-Grained Perception.}
Compared to previous IDC datasets~\cite{jhamtani2018learning,park2019robust,tan2019expressing,forbes2019neural}, our DiffGround requires a finer-grained understanding of image content. 
While IDC focuses on generating global descriptive captions of differences, DiffGround demands precise outputs with exact bounding boxes of target region. Moreover, DiffGround’s broader category set calls for stronger spatial perceiving capabilities.

\begin{figure}[htbp]
    \centering
    \includegraphics[width=0.49\textwidth]{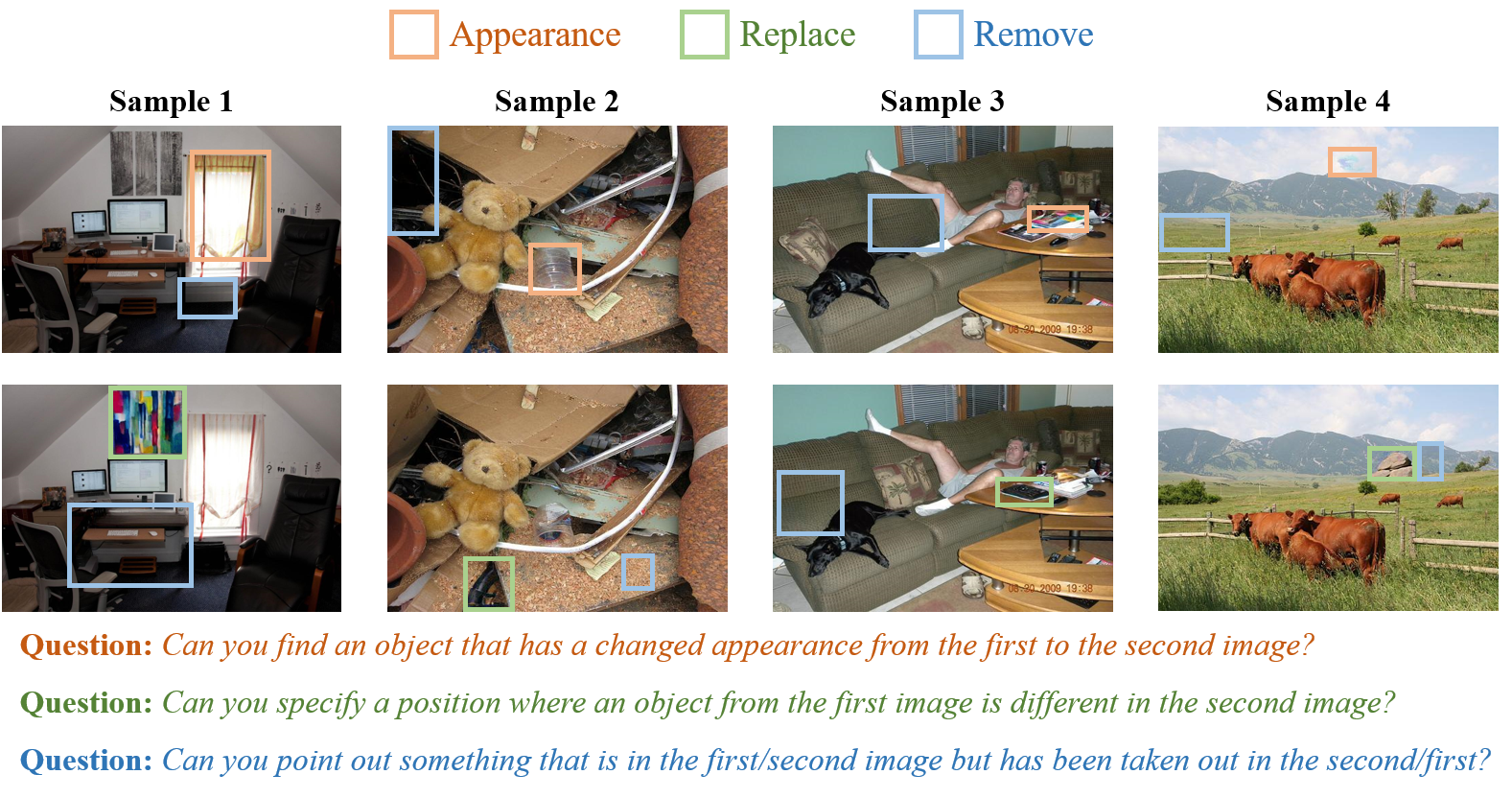}
    \vspace{-20pt}
    \caption{Visualizations of samples from our \ourdataset dataset.}
    \label{fig_data_examples}
    \vspace{-10pt}
\end{figure}

\textbf{Comprehensive Evaluation.}
As shown in \cref{fig_data_examples}, DiffGround further challenges models to capture differences at multiple levels, \ie, object category variations (\textit{replace}), object appearance discrepancies (\textit{appearance}), and the object's presence or absence (\textit{remove}). This multi-layered task design allows DiffGround to provide a more comprehensive measure of model’s ability to localize image differences.

\vspace{-10pt}

For data partition, we split 3k images, corresponding to 12k text queries, as the testing set of DiffGround, with the remaining data as training set. 
To ensure data quality, we further apply human checks on the testing set.
For the evaluation on \ourdataset, we follow the setting of previous detection-oriented benchmarks, adopting average precision with a 0.5 IoU threshold (AP@50) on bounding boxes as the evaluation metric. 
Besides, to thoroughly evaluate the model ability of grounding various difference patterns, 
we split the testing set into three subsets, corresponding to fine-grained appearance change, coarse-grained object removal and object replacement. 
AP@50 is respectively evaluated across the three above subsets and the overall testing sets, which in the following tables are represented as Diff-App, Diff-Rem, Diff-Rep, and Diff-All.

\begin{figure*}[htbp]
    \centering
    \includegraphics[width=0.86\textwidth]{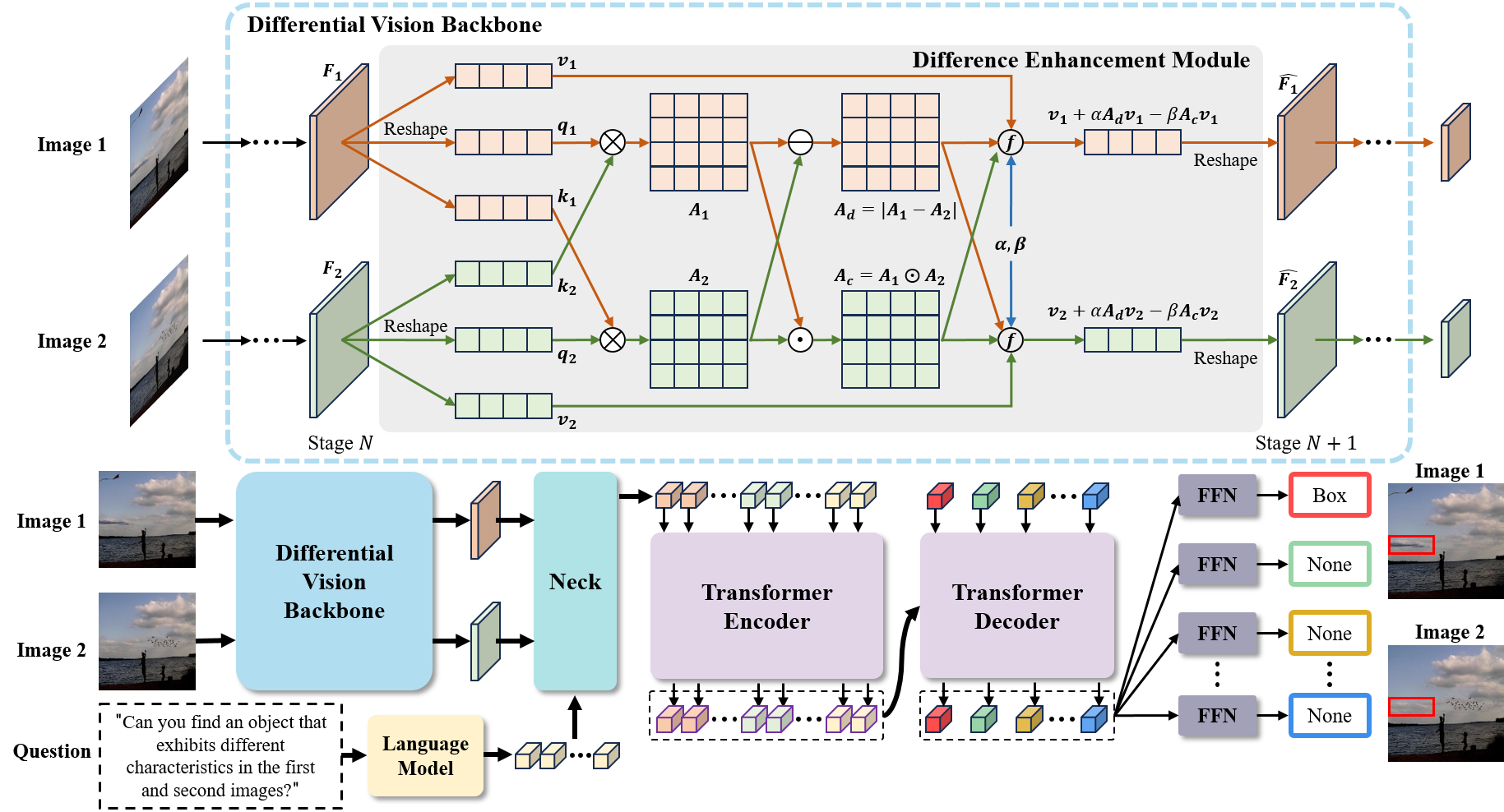}
    \vspace{-5pt}
    \caption{
    The architecture of our \ourmodel as a simple baseline for the proposed IDG task. 
    \ourmodel mainly comprises three parts: the visual and textual backbone for feature extraction, the intermediate V-L neck for cross-modal feature aggregation, and the subsequent Transformer-based encoder-decoder for multimodal feature fusion and the generation of grounding bounding boxes.
    }
    \label{fig_model}
    \vspace{-8pt}
\end{figure*}

\section{IDG Baseline DiffTracker} 
\label{sec:baseline}

Next, we describe our model \ourmodel. 
\textit{Since our original intention is to establish a simple and easy-to-follow baseline for our IDG task, the structure of \ourmodel is designed to be simple and clean}, which will be elaborated below.

\noindent \textbf{Overall Structure.} 
As shown in Fig. \ref{fig_model}, the architecture of DiffTracker is designed for precise cross-image difference grounding through structured feature extraction and interaction. 
DiffTracker first takes two similar images as input and processes them through the Differential Vision Backbone (DVB), which extracts foundational visual features from each image. 
Within this backbone, our proposed Difference Enhancement Module (DEM) is applied at each backbone stage to emphasize unique cross-image features while suppressing shared elements, helping the model isolate critical differences and enabling clearer detection of distinct visual changes.
The output from DVB is then passed to an intermediate V-L Neck, which facilitates cross-image and cross-modal interactions. 
Here, the model integrates visual features from both images with textual cues provided by a language model processing the input instruction. 
This Neck part aligns language guidance with visual discrepancies, ensuring that multimodal representations are contextually synchronized for precise grounding.
Finally, these fused representations are passed to the Transformer-based Encoder-Decoder (\ie, DETR \cite{carion2020end}), which performs targeted feature fusion and generates grounding bounding boxes around the detected differences. 
By leveraging both enhanced visual features and textual information, this Transformer Decoder accurately localizes regions of interest, completing the cross-image difference grounding task.

\noindent \textbf{Differential Vision Backbone.}
The DVB in our framework is designed to enhance cross-image differences while suppressing common features, enabling fine-grained and precise image difference grounding. 
Based on a basic vision backbone (\eg, ResNet \cite{he2016deep} or Vision Transformer \cite{dosovitskiy2020image}), we incorporate the introduced DEM to strengthen the backbone’s ability to perceive differential features between images. 
At each stage \( N \) within the vision backbone, feature maps \( F_1 \) and \( F_2 \) within this stage are hierarchically extracted from the two initial input images and reshaped to ensure dimension alignment for subsequent differential enhancement.
For each reshaped feature map, query, key, and value vectors are generated (\( q_1, k_1, v_1 \) for Image 1 and \( q_2, k_2, v_2 \) for Image 2), allowing the model to attend to specific regions and compare features between images.
Our DEM then computes a visual difference attention map \( A_d \) and a commonality attention map \( A_c \) to highlight unique features while suppressing shared elements. 
The difference and common attention maps are calculated with intermediate representations \( A_1 = q_1 \cdot k_2 \) and \( A_2 = q_2 \cdot k_1 \), where the dot product captures the cross-image attention. 
In this way, the obtained \( A_d = |A_1 - A_2| \) will capture the absolute differences between the two images, which highlight areas of significant variation, while the \( A_c = A_1 \odot A_2 \) retains similar regions.

To control the differential enhancement intensity, our DEM introduces two weight factors, \( \alpha \) and \( \beta \), which adjust the emphasis on unique and suppression on common features. The enhanced features \( \tilde{F_1} \) and \( \tilde{F_2} \) are computed as:
\begin{equation} 
    \tilde{F_1} = v_1 + \alpha A_d v_1 - \beta A_c v_1
\end{equation}
\begin{equation}
    \tilde{F_2} = v_2 + \alpha A_d v_2 - \beta A_c v_2
\end{equation}
where \( +\alpha A_d v \) amplifies unique differences, and \( -\beta A_c v \) suppresses common features, refining the model’s sensitivity to cross-image variations. 
The updated representations are then passed to the next stage of vision backbone, \( N+1 \), allowing DEM to iteratively refine features across multiple stages, establishing a robust foundation for the IDG task.

\noindent \textbf{Transformer-Based Encoder-Decoder.} 
In DiffTracker, the Transformer Decoder is a DETR-inspired V-L module specifically adapted for the IDG task, enabling fine-grained grounding of cross-image differences. 
This process is facilitated by the former Transformer-based Encoder, which first processes the multimodal features to enrich and align the extracted representations before passing them to the decoder.
The Transformer Decoder starts with a set of learned object queries that act as flexible slots for capturing potential differences. 
Each query attends to reshaped multimodal feature embeddings, combining enhanced visual features from both images with contextual cues from the input text, allowing the model to focus precisely on regions relevant to the described differences.
Through cross-attention operations, each query interacts with the encoded V-L features, progressively refining its focus on unique, stage-enhanced differential features produced by the DEM. 
Then each query is then passed through a feed-forward network to predict bounding box coordinates and class labels, either identifying detected differences or marking them as "no object" if no match is found.
Following DETR \cite{carion2020end}, our decoder employs a Hungarian matching loss for optimization, aligning predicted boxes with ground-truth differences using a combination of L1 loss and Generalized IoU \cite{rezatofighi2019generalized} for bounding accuracy, along with cross-entropy for classification.

\section{Experiments} 
\label{sec:experiment}

\subsection{Implementation Details}

Our work is implemented based on Pytorch \cite{paszke2019pytorch} and trained with 8 NVIDIA A100 GPUs. 
To enhance scalability and ease of reproducibility, we employ RoBERTa \cite{liu2019robertarobustlyoptimizedbert} as the language model and ResNet-101 \cite{he2016deep} as the basic vision backbone.
All experiments on our \ourmodel are trained with a batch size of 64 for 5 epochs, using AdamW as the optimizer with a learning rate of $5e^{-5}$ and a weight decay of $1e^{-4}$. 
During training, the images are resized while maintaining their original aspect ratio. 
The shorter side is randomly scaled between 480 and 576 (in increments of 32), and the longer side is capped at a maximum of 768. 
For fine-tuning settings of MDETR \cite{kamath2021mdetr} and Qwen-VL \cite{bai2023qwen} on \ourdataset, we adhere to their default configurations. 
For classic VG method like MDETR, which originally can perceive only a single image, we sum the features of the input image pairs extracted from a weight-shared visual encoder and pass the combined features to the following decoder part.

\subsection{Main Results and Analysis}
\label{subsec:main_results}

To thoroughly evaluate the performance of previous SOTA VG methods and our proposed DiffTracker on the newly introduced IDG task, we conducted experiments on the DiffGround testing set, categorizing included models into zero-shot and fine-tuning settings. 
Results in Table \ref{tab:sota} show that classic VG models, under zero-shot setting, struggle with interpreting user instructions and localizing subtle cross-image differences, resulting in very low IDG performance. 
Even powerful generalist models like Qwen-VL \cite{bai2023qwen} and Qwen2-VL \cite{wang2024qwen2}, despite their enhanced language comprehension and reasoning capabilities, fall short in effectively handling complex IDG tasks, especially when it comes to detecting nuanced object appearance modifications and replacements.
The substantially higher complexity of cross-image and user-intent perception in our IDG task, compared to traditional single-image VG settings, results in generally lower grounding accuracy across all models. This highlights a critical need for further research and specialized datasets in fine-grained image difference grounding, where current SOTA methods lack sufficient capability.

\vspace{-5pt}
\begin{table}[thbp]
    \small
    \setlength{\belowcaptionskip}{1.0pt}
    \begin{center}
    \caption{Comparisons with previous SOTA approaches for classic VG task on our \ourdataset testing set. 
    }
    \vspace{-2mm}
    \setlength{\tabcolsep}{1.0mm}{
    \begin{tabular}{l|cccc}
    \specialrule{.1em}{.05em}{.05em}
        \multicolumn{1}{l|}{\multirow{2}{*}{Methods}} & \multicolumn{4}{c}{\multirow{1}{*}{Image Difference Grounding}} \\ 
        \cline{2-5}
         & Diff-App & Diff-Rem & Diff-Rep & Diff-All \\
        \specialrule{.1em}{.05em}{.05em}
        \multicolumn{1}{l|}{\textit{Zero-Shot Setting}} & \multicolumn{4}{l}{} \\
        \midrule
        MDETR \cite{kamath2021mdetr} & 3.7 & 0.9 & 3.2 & 2.3 \\
        SeqTR \cite{zhu2022seqtr} & 3.4 & 1.5 & 3.9 & 2.5 \\
        Shikra~\cite{chen2023shikra} & 3.2 & 2.9 & 7.8 & 4.2 \\
        Qwen-VL~\cite{bai2023qwen} & 3.5 & 3.3 & 8.5 & 4.6 \\
        Qwen2-VL~\cite{wang2024qwen2} & 7.9 & 9.0 & 12.5 & 9.5\\
        \midrule
        \multicolumn{1}{l|}{\textit{Fine-Tuning Setting}} & \multicolumn{4}{l}{} \\
        \midrule
        MDETR \cite{kamath2021mdetr} & 9.6 & 19.1 & 7.1 & 13.7 \\
        Qwen-VL~\cite{bai2023qwen} & 35.6 & 46.4 & 31.3 & 39.9 \\
        \rowcolor{mygray} \ourmodel (Ours) & \textbf{55.5}  &  \textbf{82.6} &  \textbf{41.6} & \textbf{65.4} \\
        \specialrule{.1em}{.05em}{.05em}
    \end{tabular}
    \label{tab:sota}}
    \end{center}
    \vspace{-20pt}
\end{table}

In contrast, DiffTracker, fine-tuned on our high-quality DiffGround dataset, achieves significant performance gains across all difference patterns, underscoring the effectiveness of specialized training data in addressing fine-grained cross-image difference grounding. 
Moreover, fine-tuning on \ourdataset enables classic VG models like MDETR and Qwen-VL to make notable improvements (↑ nearly 20-40\%), bridging the performance gap in IDG tasks that conventional VG datasets cannot address. 
The above findings confirm that \ourdataset is essential for enhancing models’ abilities to perceive and localize subtle visual discrepancies across images, thereby extending the capabilities of existing VG models in complex, real-world applications.
The related qualitative analysis can be found in \textcolor{blue}{Appendix}.

\subsection{Ablation Study}

To justify the efficacy of our \ourdataset's data and \ourmodel, we conduct extensive ablation experiments on \ourdataset testing set. 
As illustrated in Sec. \ref{subsec:dataset_details}, the tables below involve the evaluations settings of each single difference pattern and across all difference patterns.

\vspace{-5pt}
\begin{table}[htbp]
    \small
    \setlength{\belowcaptionskip}{1.0pt}
    \begin{center}
    \caption{Ablation study on the data scale of \ourdataset dataset.}
     \vspace{-2mm}
     \setlength{\tabcolsep}{2.0mm}{\begin{tabular}{c|ccc|c}
      \specialrule{.1em}{.05em}{.05em} 
      \multirow{2}{*}{Ratio} & \multicolumn{4}{c}{DiffGround}\\
       \cline{2-5}
        & Diff-App & Diff-Rem & Diff-Rep & Diff-All  \\
        \hline
         25\% & 10.1 & 39.2 & 16.5 & 26.2 \\ 
         50\% &  37.7 & 66.1 & 26.2 & 48.8 \\ 
         75\% & 50.6 & 78.2 & 38.6 & 61.5 \\ 
        \rowcolor{mygray} 100\% & \textbf{55.5}  &  \textbf{82.6} &  \textbf{41.6} & \textbf{65.4}  \\
    \specialrule{.1em}{.05em}{.05em}
    \end{tabular}
    \vspace{-30pt}
    \label{tab:Data_Scale}}
    \end{center}
\end{table}

\paragraph{Effect of Data Scale.}
First, we investigate the impact of using varying proportions of the DiffGround dataset on model performance, as shown in Table \ref{tab:Data_Scale}. 
It is evident that as the proportion of training samples increases, the model’s accuracy improves steadily across all difference pattern settings, validating the effectiveness and high quality of our data. 
Training on the full dataset yields a Diff-All accuracy of 65.4\%, significantly outperforming the 26.2\% and 48.8\% accuracies achieved with 25\% and 50\% of the data, respectively. 
This upward trend highlights the benefits of DiffGround, as the model consistently gains from larger training sets without signs of diminishing returns, suggesting strong potential for further enhancing fine-grained cross-image difference grounding performance as the dataset scales up.

\vspace{-5pt}
\begin{table}[htbp]
    \small
    \setlength{\belowcaptionskip}{1.0pt}
    \begin{center}
    \caption{Ablation study on the difference patterns of \ourdataset data. Appearance, Remove and Replace denote employing IDG data of corresponding difference pattern for training.}
     \vspace{-3mm}
     \setlength{\tabcolsep}{0.2mm}{\begin{tabular}{c|c|c|ccc|c}
      \specialrule{.1em}{.05em}{.05em} 
      \multirow{2}{*}{Appearance} & \multirow{2}{*}{Remove} & \multirow{2}{*}{Replace} &  \multicolumn{4}{c}{DiffGround}\\
      \cline{4-7}
         &   &   & Diff-App & Diff-Rem & Diff-Rep & Diff-All  \\
        \hline
        & \ding{51}  & \ding{51}  &  2.8 &  78.3 &  33.7 &  48.2\\
        \ding{51} &  & \ding{51}  &  28.3 & 2.4 & 22.4  & 13.9 \\
        \ding{51} & \ding{51} &   & 54.5  & 80.9  &   2.5 &  54.6\\
        \rowcolor{mygray} \ding{51} & \ding{51} & \ding{51} & \textbf{55.5}  &  \textbf{82.6} &  \textbf{41.6} & \textbf{65.4}   \\
    \specialrule{.1em}{.05em}{.05em}
    \end{tabular}
    \vspace{-30pt}
    \label{tab:Data_Pattern}}
    \end{center}
\end{table}

\paragraph{Effect of Visual Difference Pattern.}
We further investigate the impact of training data from various visual difference patterns on IDG task performance. 
As shown in Table \ref{tab:Data_Pattern}, removing data related to appearance variance, removal, or replacement patterns leads to a noticeable decline in model performance within the corresponding pattern category. 
Moreover, excluding any single pattern significantly reduces overall IDG accuracy, highlighting the importance of each type of visual difference data. 
Notably, the model’s lowest accuracy occurs when removal pattern data is excluded, indicating the critical role this pattern plays in effective IDG. 
Conversely, the highest overall accuracy is achieved when all patterns are included, underscoring the value of a diverse training set that encompasses all difference types and the necessity of varied visual patterns to ensure robust IDG performance across scenarios.

\vspace{-5pt}
\begin{table}[htbp]
    \small
    \setlength{\belowcaptionskip}{1.0pt}
    \begin{center}
    \caption{Ablation study on the insertion stage of our DEM.}
     \vspace{-3mm}
     \setlength{\tabcolsep}{1.4mm}{\begin{tabular}{c|c|c|c|ccc|c}
      \specialrule{.1em}{.05em}{.05em} 
      \multicolumn{4}{c|}{Stage} &  \multicolumn{4}{c}{DiffGround}\\
      \cline{1-8}
         0 & 1 & 2 & 3 & Diff-App & Diff-Rem & Diff-Rep & Diff-All  \\
        \hline
         &   &   &   &  28.7 & 31.5 & 23.2  & 28.7 \\ 
        \ding{51} &   &   &   &  50.4 &  71.2 &36.5 & 57.4 \\
        \ding{51} & \ding{51} &   &   & 54.5 & 80.6 & 40.8  &    64.2\\
        \ding{51} & \ding{51} & \ding{51} &    &    54.0&  82.4 &   40.6 &    64.9\\
        \rowcolor{mygray} \ding{51} & \ding{51} & \ding{51} & \ding{51} & \textbf{55.5}  &  \textbf{82.6} &  \textbf{41.6} & \textbf{65.4}   \\
    \specialrule{.1em}{.05em}{.05em}
    \end{tabular}
    \vspace{-30pt}
    \label{tab:insertion_stage}}
    \end{center}
\end{table}

\paragraph{Effect of Differential Enhancement Stage.}
We further explore the effect of inserting our designed DEM at different stages within vision backbone. 
The DEM is designed to amplify differential features while suppressing commonalities between image pairs, thus improving the model's ability to localize cross-image differences. 
As shown in Table \ref{tab:insertion_stage}, the baseline performs poorly on IDG task without our DEM, indicating limited capacity to capture fine-grained visual differences.
As our DEM is progressively inserted at various stages of the hierarchical vision backbone, we observe consistent improvements on IDG performance, with the best results achieved when applied across all stages. 
Notably, inserting the DEM at earlier stages yields larger performance gains than later-stage insertions, allowing the model to identify visual differences earlier and progressively refine these differential representations. 
This multi-stage application of our DEM designing incrementally enhances the model’s focus on unique visual discrepancies, greatly boosting its effectiveness on the IDG task.

\vspace{-5pt}
\begin{table}[htbp]
    \small
    \setlength{\belowcaptionskip}{1.0pt}
    \begin{center}
    \caption{Ablation study on the intensity weight $\alpha$, $\beta$ in our DEM.}
     \vspace{-2mm}
     \setlength{\tabcolsep}{2.0mm}{\begin{tabular}{c|ccc|c}
      \specialrule{.1em}{.05em}{.05em} 
      \multirow{2}{*}{$\alpha$, $\beta$} & \multicolumn{4}{c}{DiffGround}\\
       \cline{2-5}
        & Diff-App & Diff-Rem & Diff-Rep & Diff-All  \\
        \hline
         0.25 &  46.7 &  71.9 &  30.9 &  55.3 \\ 
         0.5 &  54.4 &  81.1 & 41.1  &  64.5 \\ 
         \rowcolor{mygray} 0.75 & \textbf{55.5}  &  \textbf{82.6} &  \textbf{41.6} & \textbf{65.4}  \\
         1.0 &  55.0 &  80.7 &  40.4 &  64.1 \\ 
    \specialrule{.1em}{.05em}{.05em}
    \end{tabular}
    \vspace{-30pt}
    \label{tab:intensity_alpha}}
    \end{center}
\end{table}

\paragraph{Effect of Differential Enhancement Intensity.}
Finally, we exploit the effect of varying the intensity weights \( \alpha \) and \( \beta \) in our DEM for the IDG task. 
These two weights jointly control the enhancement of differential features and suppression of common features between the input images. 
As illustrated in Table \ref{tab:intensity_alpha}, the model achieves optimal performance at an intermediate intensity level of \( \alpha, \beta = 0.75 \). 
As \( \alpha \) and \( \beta \) increase from 0.25 to 0.75, the model’s grounding accuracy steadily improves, indicating an enhanced ability to identify and localize fine-grained visual discrepancies.
However, with too high intensities, model performance slightly declines. 
We believe this is because lower intensities provide insufficient enhancement, leading to weak differential feature emphasis, whereas moderate intensities effectively boost the model's grounding accuracy. 
Beyond the sweet spot of \( \alpha, \beta = 0.75 \), excessive enhancement may cause the model to overemphasize differences, neglecting essential contextual information and impairing overall performance. 
These results highlight the importance of balanced differential enhancement in our DEM structure, ensuring an ideal trade-off that maximizes IDG task performance.

\section{Conclusion, Broader Impact and Limitation} 
\label{sec:conclusion}

In this work, we introduce Image Difference Grounding (\ourtask), a pioneering V-L task aimed at accurately identifying and localizing visual differences between paired images based on natural language. 
We build \ourdataset, a large-scale, high-quality dataset containing diverse image variations, each annotated with textual queries highlighting fine-grained differences. 
We propose \ourmodel, a simple yet effective framework that leverages our designed Difference Enhancement Module to stage-wise highlight visual differences while suppressing common features between paired images. 
Through a DETR-based Transformer decoder, \ourmodel can accurately localize the cross-image discrepancies with specific instructions.
Experimental results show that \ourmodel significantly outperforms traditional VG models on \ourdataset benchmark, demonstrating its superior ability to detect subtle visual changes with linguistic cues.
By releasing \ourdataset and \ourmodel, we aim to provide key resources that advance nuanced V-L research and foster the development of multimodal intelligent systems for real-world human-machine interactions.

While this work advances the classic VG task towards fine-grained cross-image difference understanding, there are still limitations that present opportunities for future exploration. 
One potential limitation is the current scale of our \ourdataset and \ourmodel could be further expanded to boost IDG performance. 
Moreover, while \ourmodel primarily focuses on visual difference grounding, it currently lacks the ability to generate textual responses for seamless human-computer interactions. 
However, integrating our framework with advanced large language models could address this gap, enabling fluid textual response generation alongside difference localization. 
This opens a promising avenue for future research, focused on developing a more versatile framework that integrates fine-grained grounding with language generation. 
By exploring the synergies between IDG and IDC, we aim to unify these tasks into a cohesive system. This would enhance the agents' capability to understand and respond to complex real-world scenarios, contributing to the development of more intelligent multimodal systems.

\clearpage

\appendix
\section{Appendix}

In this appendix section, we provide following items:
\begin{itemize}[noitemsep,leftmargin=*]
    \item (Sec. \textcolor{red}{1}) More prompt details for Step 1: Object Selection and Filtering in our data collection engine. 
    \item (Sec. \textcolor{red}{2}) More prompt details for Step 2: Instruction and Query Generation in our data collection engine. 
    \item (Sec. \textcolor{red}{3}) Visual comparison between previous grounding state-of-the-art methods and our proposed DiffTracker. 
\end{itemize}

\section{More Details about Step 1's Prompt}
In Step 1, we use Qwen2-7B~\cite{yang2024qwen2} as the language model to score the labels. The model is provided with the prompt listed below as input. For each object label, the model outputs whether the object is isolated, whether it is part of another object, and whether it is in the foreground. 
\begin{tcolorbox}[title=Isolation Discriminate Prompt]
\renewcommand\baselinestretch{1.5}\selectfont
\small

Given the object label below, answer the question.

Is the object isolated, or is it tightly connected to other objects in natural environment?

If it is isolated, return YES, else, return NO

Object Label: \{\}

Response:
\end{tcolorbox}

\begin{tcolorbox}[title=Part Discriminate Prompt]
\renewcommand\baselinestretch{1.5}\selectfont
\small

Given the object label below, answer the question.

Is the object a standalone item or a part of a larger object?

If it is a part of a larger object, return YES, else, return NO

Object Label: \{\}

Response:
\end{tcolorbox}

\begin{tcolorbox}[title=Foreground Discriminate Prompt]
\renewcommand\baselinestretch{1.5}\selectfont
\small
Given the object label below, answer the question.

Is the object a foreground item or a background item?

If it is a background item, return YES, else, return NO

Object Label: \{\}

Response:
\end{tcolorbox}

Based on these three characteristics, the model then generates a final score to determine whether the object should be filtered or removed.

\section{More Details about Step 2's Prompt}

We first use Qwen2-VL~\cite{wang2024qwen2} to extract the target's attributes, including 'color', 'shape', 'texture', 'material', and 'pattern'. The following prompt is used along with the input image to obtain the corresponding attributes. If the model returns 'unknown' for any attribute, that attribute will be excluded.

\begin{tcolorbox}[title=Attribute Prompt]
\renewcommand\baselinestretch{1.5}\selectfont
\small
What is the \{attribute\} of \{label\} in this image? If you can not decided the answer, return unknown. Give a short word or phrase as the answer.
\end{tcolorbox}

For objects that require object-level replacement, we use the following prompt to generate the replacement object category, with Qwen2-7B~\cite{yang2024qwen2} as the generator.

\begin{tcolorbox}[title=Object Replace Instruction Prompt]
\renewcommand\baselinestretch{1.5}\selectfont
\small
Given the label of an object, provide a new label that falls under the different broad category. 

Ensure that the two objects are not similar in appearance or sound.

Ensure that there is a significant visual difference between the object labels before and after.

Input Format:

Object Label: [insert object]

Output Format:

New Object Label: [output new object]

Example:

Object Label: Apple

Expected Output:

New Object Label: Bowl

Input:

Object Label: \{\}

Output:

New Object Label: 
\end{tcolorbox}

For objects that require attribute-level replacement, we randomly select one of the attributes obtained earlier and use the following prompt to generate the replacement attribute, with Qwen2-7B~\cite{yang2024qwen2} as the generator.

\begin{tcolorbox}[title=Attribute Replace Instruction Prompt]
\renewcommand\baselinestretch{1.5}\selectfont
\small
Given an object label and an attribute, generate a new attribute that belongs to the same category as the original attribute but represents a different specific value.

Ensure that there is a significant visual difference between the attributes before and after.

Input Format:

Object Label: [insert object]

Original Attribute: [insert attribute]

Output Format:

New Attribute: [output new attribute]

Example:

Object Label: Apple

Original Attribute: Red

Expected Output:

New Attribute: Yellow

Input:

Object Label: \{\}

Original Attribute: \{\}

Output:

New Attribute: 
\end{tcolorbox}

\begin{figure}[thbp]
    \centering
    \includegraphics[width=0.48\textwidth]{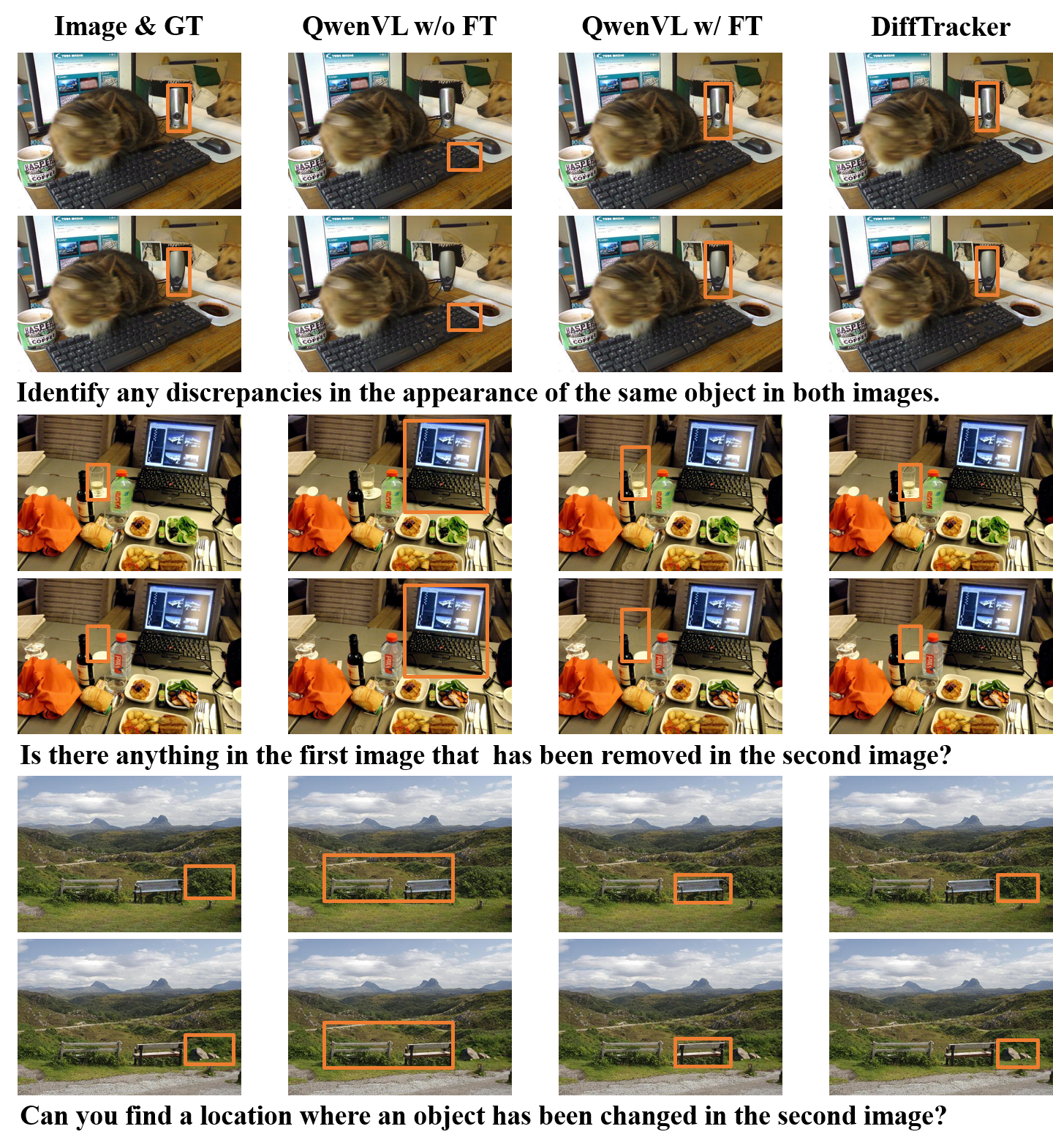}
    \vspace{-5pt}
    \caption{The visual comparison of image difference grounding results on our DiffGround test set. (a) the input image and corresponding ground truth. (b) Qwen-VL. (c) Qwen-VL fine-tuned with our DiffGround data. (d) our DiffTracker.}
    \label{fig_SOTAComparison_vis}
    \vspace{-10pt}
\end{figure}

\section{Visual Comparison for Qualitative Analysis}

To validate the effectiveness of our DiffGround dataset and the DiffTracker model for the image difference grounding task, we conduct the qualitative comparison between the classic state-of-the-art visual grounding method Qwen-VL \cite{bai2023qwen}, both before and after fine-tuning on our DiffGround dataset, and our DiffTracker model. The visualization results presented in the second and third columns of Fig. \ref{fig_SOTAComparison_vis} clearly demonstrate that our constructed DiffGround dataset significantly helps the previous grounding SOTA method enhance its ability to perceive subtle visual differences between two images at a fine-grained level. Furthermore, as shown in the third and fourth rows of Fig. \ref{fig_SOTAComparison_vis}, our DiffTracker with its targeted design for visual difference feature perception outperforms the fine-tuned Qwen-VL, which is built upon large language model, showing a clear and consistent advantage.

\clearpage

{
    \small
    \bibliographystyle{ieeenat_fullname}
    \bibliography{main}
}

\end{document}